\documentclass[sigconf]{acmart}

\usepackage{algorithm}  
\usepackage{algpseudocode}  

  \makeatletter
  \newcommand\fs@ruled@notop{%
    \def\@fs@cfont{\bfseries}\let\@fs@capt\floatc@ruled
    \def\@fs@pre{}
    \def\@fs@post{\kern2pt\hrule\relax}
    \def\@fs@mid{\kern12pt\hrule\kern2pt}  
    \let\@fs@iftopcapt\iftrue}
  \renewcommand\fst@algorithm{\fs@ruled@notop}
  \makeatother

\usepackage{booktabs}  

\usepackage[capitalise]{cleveref}  

\usepackage{paralist}  

\usepackage[font = scriptsize]{subfig}  

\usepackage{tabularx}  

\newcommand{\eg}{e.\,g.}
\newcommand{\ie}{i.\,e.}

\graphicspath{{figures/}}

\renewcommand\footnotetextcopyrightpermission[1]{}

\begin{document}
  

\title[ALF]{ALF -- A Fitness-Based Artificial Life Form for Evolving Large-Scale Neural Networks}

\author{%
  Rune Krauss \texorpdfstring{\quad}{} Marcel Merten \texorpdfstring{\quad}{} Mirco Bockholt \texorpdfstring{\quad}{} Rolf Drechsler \texorpdfstring{\\}{}\texorpdfstring{\vspace{1ex}}{}
  Group of Computer Architecture, University of Bremen, Germany \texorpdfstring{\\}{}
  Cyber-Physical Systems, DFKI GmbH, Germany \texorpdfstring{\\}{}
  \{krauss, mar\_mer, bockholt, drechsler\}@uni-bremen.de
}

\begin{abstract}
  \textit{Machine Learning}~(ML) is becoming increasingly important in daily life. In this context, \textit{Artificial Neural Networks}~(ANNs) are a popular approach within ML methods to realize an artificial intelligence. Usually, the topology of ANNs is predetermined. However, there are problems where it is difficult to find a suitable topology. Therefore, \textit{Topology and Weight Evolving Artificial Neural Network}~(TWEANN) algorithms have been developed that can find ANN topologies and weights using genetic algorithms. A well-known downside for large-scale problems is that TWEANN algorithms often evolve inefficient ANNs and require long runtimes.

  To address this issue, we propose a new TWEANN algorithm called \textit{Artificial Life Form}~(ALF) with the following technical advancements:
  \begin{inparaenum}
    \item speciation via structural and semantic similarity to form better
          candidate solutions,
	  \item dynamic adaptation of the observed candidate solutions for better 
          convergence properties, and
	  \item integration of solution quality into genetic reproduction to
          increase the probability of optimization success.
  \end{inparaenum}
  Experiments on large-scale ML problems confirm that these approaches allow the fast solving of these problems and lead to efficient evolved ANNs.
\end{abstract}

\thanks{
  This work is licensed under the \href{https://arxiv.org/licenses/nonexclusive-distrib/1.0/license.html}{arXiv.org perpetual, non-exclusive license}. \\
  \copyright{} 2021 Copyright held by the authors.
}

\maketitle



\section{Introduction}
\label{sec:introduction}

In daily life, the usage of \emph{Machine Learning}~(ML) is becoming more popular, especially \emph{Reinforcement Learning}~(RL)~\cite{mosavi2020comprehensive}. As an important ML method, RL improves the decision-making ability of an intelligent agent by interacting with an unknown environment, and has therefore attracted more attention in economics. Furthermore, there are great application perspectives in domains such as robotics~\cite{kober2013reinforcement} or human-level intelligent game play~\cite{torrado2018deep}.

Within ML methods, artificial intelligences are often realized by \emph{Artificial Neural Networks}~(ANNs)~\cite{nguyen2019applications, liu2014recursive, graves2014online, zhang2018recommendation}. The topology of ANNs is application-specific and usually chosen based on empirical evidences. Unfortunately, although two ANNs with different topologies can theoretically represent the same function~\cite{lewicki2004approximation}, the most efficient topology is not readily identifiable. In particular, for sequential problems where an output depends on previous decisions, it is difficult to find such a topology.

\emph{Neuroevolution}~(NE) is the artificial evolution of ANNs using \emph{Genetic Algorithms}~(GAs) and explores through a search space for an ANN that performs well on a given task. It offers an alternative to statistical methods for solving control tasks that attempt to estimate the utility of actions in states of the world~\cite{arulkumaran2017brief}. Studies have shown that NE is more efficient than RL methods like Q-learning and Adaptive Heuristic Critic for several RL problems like robotic control~\cite{moriarty1997forming, stanley2003competitive}. Compared to RL, a direct exploration in the search space is possible without the requirement for indirect inferences from value functions. Furthermore, problems can be dealt with effectively where gradients are difficult to compute or do not exist.

Besides conventional NE, there are \emph{Topology and Weight Evolving Artificial Neural Network}~(TWEANN) algorithms that evolve topology in addition to ANN weights. However, a downside of TWEANN algorithms is that in order to solve large-scale RL problems, the size of ANNs can grow rapidly and lead to inefficient large ANNs~\cite{stanley2009hypercube}. Thus, the evolution of ANNs has a long runtime and large memory requirements, which makes it difficult to solve specific tasks. The reason for this is that a large search effort is made within a complex search space, where there are few suitable solutions. Although efforts have been made to improve the capability of TWEANN algorithms to solve problems with large state spaces, there is still room for improvement for existing technologies~\cite{berg2013critical, peng2018neat, whiteson2005automatic}. For example, HyperNEAT uses an indirect encoding method to evolve large ANNs, but performs poorly on problems with high-level and geometrically unrelated features~\cite{berg2013critical}.

This paper proposes a new TWEANN algorithm called \emph{Artificial Life Form}~(ALF), which has the following main advantages over traditional TWEANN algorithms:
\begin{inparaenum}
  \item speciation via structural and semantic similarity to form better
        candidate solutions,
  \item dynamic adaptation of the observed candidate solutions for better
        convergence properties, and
  \item integration of solution quality into genetic reproduction to increase
        the probability of optimization success.
\end{inparaenum}
The synergy of these components improves the runtime and memory usage, leading to efficient evolved ANNs. Experimental results on large-scale RL problems confirm this performance.

In the remainder of this paper, the made contributions are described in the following. The next section gives an overview of TWEANN and the resulting design challenge. Afterwards, \Cref{sec:alf} presents the proposed solution called ALF for these problems by describing its three components:
\begin{inparaenum}
  \item \emph{Semantic and Structural Speciation}~(SSS),
  \item \emph{Dynamic Adaptation of Population}~(DAP), and
  \item \emph{Fitness-Based Genetic Operators}~(FBGO).
\end{inparaenum}
Then, the performance of ALF on large-scale RL problems is considered in~\Cref{sec:experimentalResults}, which summarizes the conducted experimental evaluations. Finally, the paper is concluded in~\Cref{sec:conclusion}.

\section{Preliminaries}
\label{sec:preliminaries}

This section addresses issues of evolutionary computation, ANNs and biology in an attempt to keep this work self-contained.

\subsection{Genetic Algorithms~(GAs)}
\label{sec:preliminaries-gas}

GAs are a family of computational models inspired by Darwinian natural selection. The original GA was introduced by Holland~\cite{holland1992adaptation}.
\begin{algorithm}[t]
  \caption{Possible realization of a traditional GA}
  \label{alg:preliminaries-gas-traditionalGa}
  \begin{algorithmic}[1]
    \Require{Fitness threshold $f_t$}
    \Ensure{Best individual $c \in P(t)$}
    \State $t \gets 0$
    \State Initialize population $P(t)$ consisting of $n$ individuals
    \While{fitness threshold $f_t$ is not reached}
      \State Evaluate fitness $f(c) \ \forall c \in P(t)$
      \For{$i = 1, 2, \dots, n$}
        \State Select parents $c_l, c_r \ (l, r \in [1, n], l \neq r)$
        \State $c' \gets$ crossover($c_l$, $c_r$)
        \State mutate($c'$)
        \State $P(t+1)_i \gets c'$
      \EndFor
      \State $t \gets$ t + 1
    \EndWhile
    \State \Return{best $c \in P(t)$}
  \end{algorithmic}
\end{algorithm}
\Cref{alg:preliminaries-gas-traditionalGa} shows a simplified traditional realization of a GA, in which a population~$P$ consists of $n$~candidate solutions~(individuals), where each individual~$c$ represents a point in the search space.

Traditionally, the initial population consists of random individuals, whereby various strategy parameters like the population size are predetermined~(lines 1 to 2). Each population created in generation~$t$ is called $P(t)$. The objective is to explore the search space for a solution that optimizes a performance criterion~(fitness threshold)~$f_t$~(line 3). A potential solution~(individual's phenotype) to a problem is encoded in a chromosome-like data structure, \ie, a string of genes, called a genome~(genotype). Genes can be binary, real or correspond to any other searchable encoding scheme. The genotype is transformed into a phenotype by a specific decoding function and evaluated by a fitness function~$f$ for a task~(line 4). The parents with higher fitness are then allowed to mate to reproduce. They are placed in a mating pool~(line 6) and have a higher probability of reproduction. Here the hope is that the crossover of two good gene sets will result in better genes~(line 7). Mutations cause random perturbations to the genes during reproduction and enable the exploration of new regions in the search space~(line 8). Analogous to biological mutations, they ensure genetic diversity from one population to the next~(lines 9 to 11)~\cite{thede2004introduction}. Finally, the best individual is returned~(line 13).

In general, GAs can vary significantly. For example, the way a genotype is encoded can have a significant impact on the number of generations or result in no solution being found~\cite{stanley2003competitive}. In addition, it is also significant how large the search space or how large the number of solutions is. Therefore, one area where ALF differs from traditional GAs is that the number of individuals or, respectively, the number of genes, can be dynamically increased and decreased at runtime, which provides more flexibility and further reduces the odds of getting stuck in local optima.

\subsection{Artificial Neural Networks~(ANNs)}
\label{sec:preliminaries-anns}

ANNs are computational models inspired by biological neural networks that seek to mimic the behavior and adaptive capabilities of the central nervous system~\cite{grossi2008introduction}. They are theoretically capable of approximating any continuous function~\cite{lewicki2004approximation}. This property makes them powerful tools for control and prediction.

An ANN can be conceptualized in terms of layers and typically consists of fully connected processing units called neurons~(nodes). There is an input layer with input nodes that receive data from the external world and pass their activations forward to other nodes via weighted connections. Usually, the number of input nodes corresponds to the number of features in the dataset. The last layer in an ANN is the output layer, whose outputs are used by the respective system containing the ANN. Typically, the number of output nodes corresponds to the number of classes to be predicted in the dataset. There may be hidden nodes between the input and output layers, which are located in hidden layers.
\begin{figure}[t]
  \centering
  \subfloat[Feed Forward]{\includegraphics[width=0.23\textwidth]{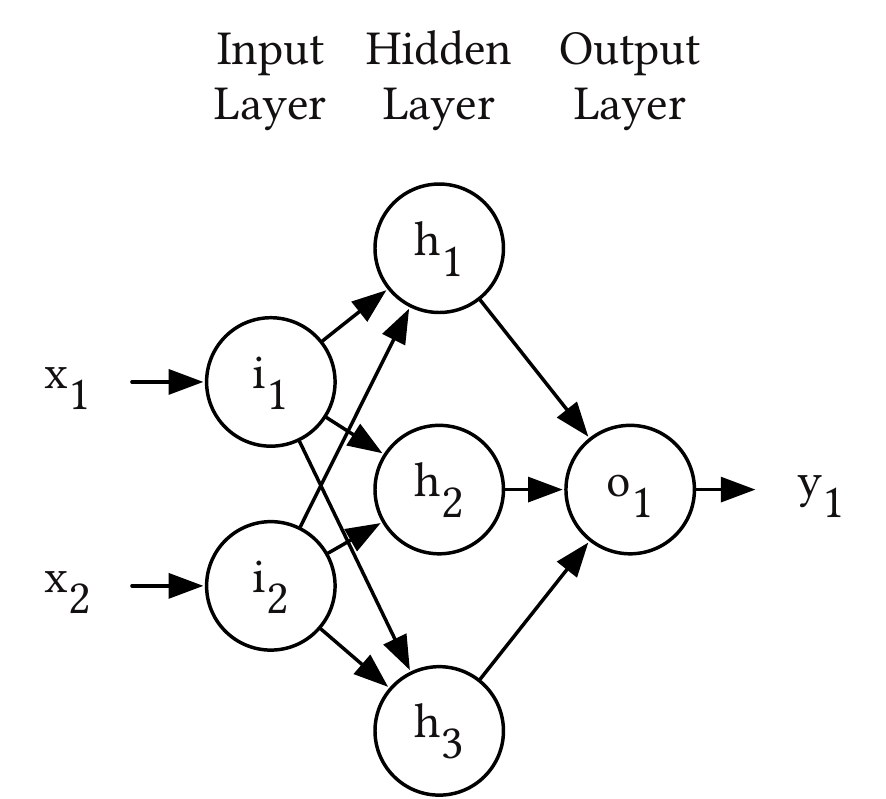}
    \label{fig:preliminaries-anns-traditionalFfn}}
  \hfil
    \subfloat[Feed Forward~(non-traditional)]{\includegraphics[width=0.23\textwidth]{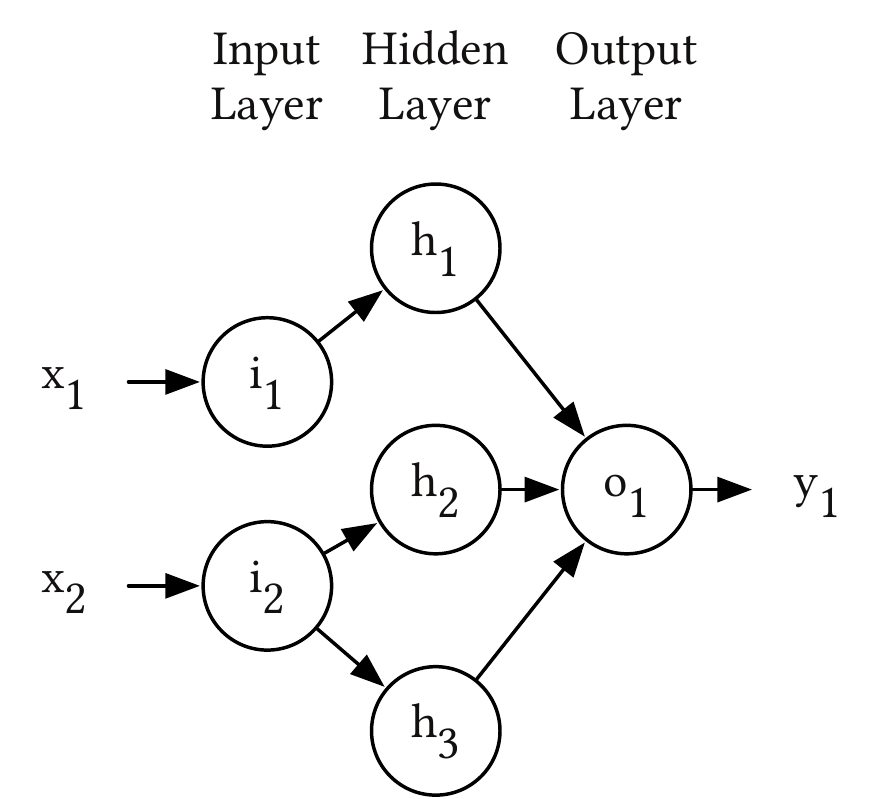}
    \label{fig:preliminaries-anns-partialFfn}}
  \caption{Fully and partially connected FFN topologies}
  \label{fig:preliminaries-anns-traditionalPartialFfns}
\end{figure}
Different types of networks are distinguished. \Cref{fig:preliminaries-anns-traditionalPartialFfns} shows ANNs, in which activation starts at the inputs and flows forwards to the outputs of the network. Such ANNs are called \emph{Feed Forward Networks}~(FFNs)~\cite{lavine2010feed}. Although ANNs are traditionally considered to be strictly layered and fully connected, which is shown in~\Cref{fig:preliminaries-anns-traditionalFfn}, ALF does not follow this tradition as its own ANN topologies can be evolved. Evolved ANN topologies are therefore unlikely to be fully connected, but rather partially connected as shown in \Cref{fig:preliminaries-anns-partialFfn}.

Each node calculates a weighted sum of its respective inputs. The sum passes through an activation function~$\phi$, which squashes the activation into a certain range. For each node $j$, the output $y_j$ for a passed input vector $x$ is given by \Cref{eq:preliminaries-anns-weightedSum}
\begin{equation}
  \label{eq:preliminaries-anns-weightedSum}
  y_j = \phi \left(\sum_i w_{ij}x_i + w_b\right)\textrm{,}
\end{equation}
where $w_{ij}$ is the connection weight from node $i$ to $j$ and $w_b$ is the connection weight for the bias. The activity level of the bias unit is usually 1 and the weight to a node can be positive or negative. If, \eg, there is a weak input from other nodes, the bias ensures that the unit remains active with a positive weight. This is useful if a node is frequently activated. The activation function is usually non-linear~\cite{olgac2011performance} and can represent, \eg, the sigmoid function $\frac{1}{1 + e^{-x}}$. In this case, the activation would be squashed into the range $[0.0, 1.0]$.
\begin{figure}[t]
  \centering
  \subfloat[Recurrent]{\includegraphics[width=0.23\textwidth]{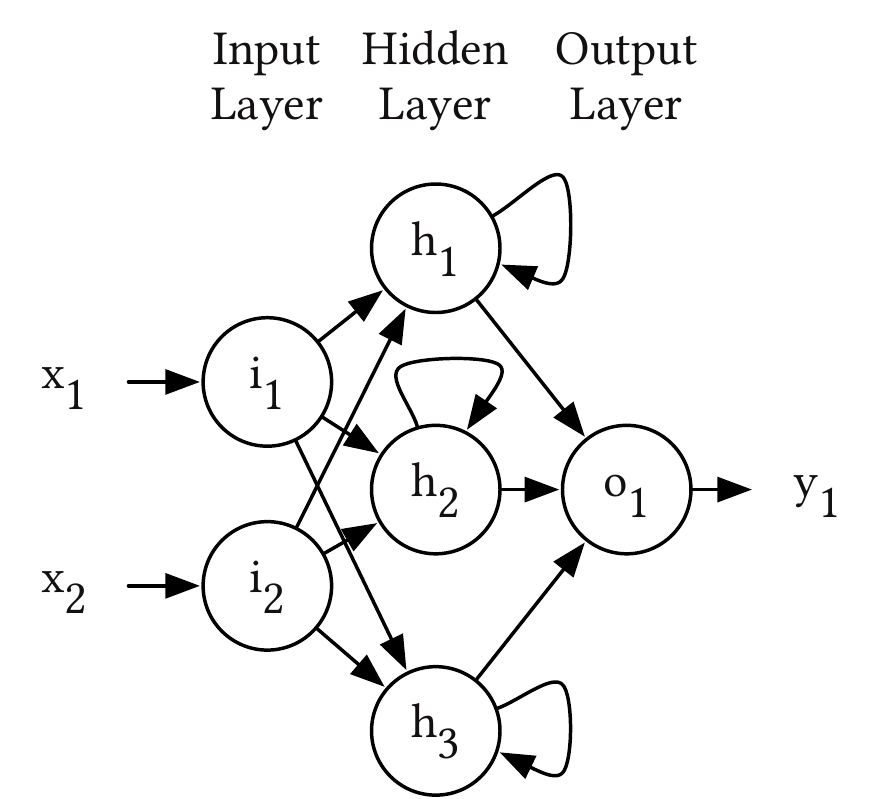}
    \label{fig:preliminaries-anns-traditionalRnn}}
  \hfil
  \subfloat[Recurrent~(non-traditional)]{\includegraphics[width=0.23\textwidth]{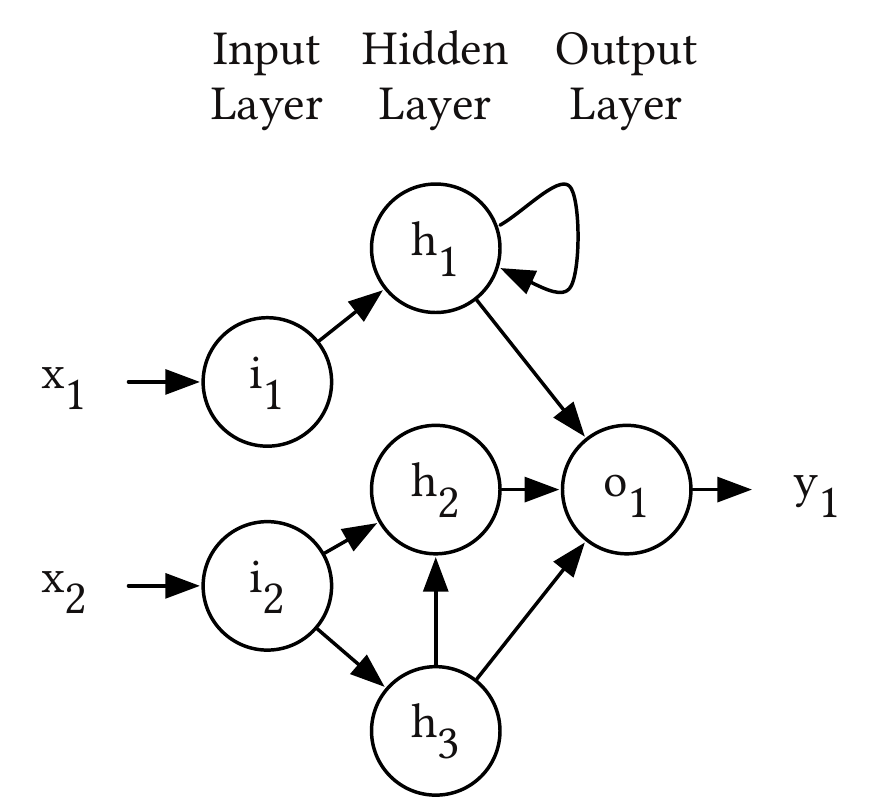}
    \label{fig:preliminaries-anns-partialRnn}}
  \caption{Fully and partially connected RNN topologies}
  \label{fig:preliminaries-anns-traditionalPartialRnns}
\end{figure}
ANNs can also have feedback connections that flow backwards instead of forwards. ANNs with feedback connections are called \emph{Recurrent Neural Networks}~(RNNs). Unlike FFNs, inputs are interdependent as part of a sequence and RNNs are therefore useful for learning temporal dependencies~\cite{du2014recurrent}. This type of network is shown in \Cref{fig:preliminaries-anns-traditionalPartialRnns}, where \Cref{fig:preliminaries-anns-traditionalRnn} represents a traditional RNN, analogous to the FFNs. ALF can also evolve recurrent structures if required, which is shown in~\Cref{fig:preliminaries-anns-partialRnn}. This is useful because it is not always necessary to have a recurrent connection for each hidden node. Often the temporal dependencies can be captured with only a few recurrent connections. Consequently, it would be a waste of search space to use unnecessary recurrent connections.

Traditionally, the ANN topology is determined a priori before training. The weights can then be trained using gradient descent methods like backpropagation~\cite{otair2006efficient}, where a loss function is minimized by adjusting the weights. As a search heuristic, GAs have some advantages compared to other search methods. While gradient descent methods can get trapped in the local minima in the error surface, GAs minimize this pitfall by sampling multiple points on the error surface~\cite{stanley2003competitive}. In particular, feedback tends to cause unexpected effects in the error surface, so that local minima are more likely to occur at suboptimal points on the surface. Additionally, training RNNs is slower and less reliable~\cite{informatik2003gradient}. In this context, the vanishing gradient problem~\cite{hochreiter1998vanishing} in relation to RNNs should be mentioned. The gradient can decrease exponentially in the number of multiplications in RNNs, so that states do not contribute to the learning process. This circumstance also applies to FFNs analogously, with the exception that RNNs usually have a deeper structure. In contrast, GAs require minimal a priori knowledge about the problem domain and different parts of the search space can easily be searched simultaneously. The GAs can even search for ANNs in sparse reinforcement. In this context, the phenotype mentioned in \Cref{sec:preliminaries-gas} is gradually optimized by GAs.

\section{Artificial Life Form~(ALF)}
\label{sec:alf}

In this work, we propose a new TWEANN algorithm called ALF, which is a solution that addresses the issue of efficiently evolving ANNs for large-scale RL problems. For this purpose, ALF integrates three main components:
\begin{inparaenum}
  \item SSS,
  \item DAP, and
  \item FBGO.
\end{inparaenum}
We begin by explaining the genetic encoding used in ALF, followed by a detailed description of the components that address the issue mentioned.

\subsection{Genetic Encoding}
\label{sec:alf-geneticEncoding}

ALF uses a direct encoding scheme designed to compare and crossover genes efficiently. Thus, a direct genotype-phenotype mapping exists for the ANN, specifying each weighted connection and layer node with an activation function that appears in the phenotype.

Genotypes are an ordered representation of genes consisting of layer connections and activation functions. A layer connection consists of a source layer id, destination layer id, and associated connection matrix. This matrix is a non-empty \mbox{$(m \times n) = (a_{ij})$} matrix with $n$ source nodes, $m$ destination nodes, and a vector~$a$.

Each floating-point number in the matrix at position \\
\mbox{$a_{ij} = [i \cdot n - n + j]$ ($1 \leq i \leq m$, $1 \leq j \leq n$)} is the weight of the respective connection between node $i$ and $j$.

In this context, we determined experimentally that this kind of implementation provides efficient cache support and optimizes memory accesses with respect to various operations such as multiplication within the ANN evaluation.

In ALF, a bias unit can be configured. The bias unit is an input that is always set to 1. If there is a connection from a bias unit to a layer, it is encoded into the connection matrices of the respective layer, whereby connections to the input layer are prohibited.

\begin{figure}[t]
  \centering
  \subfloat[Genotype]{\includegraphics[width=0.32\textwidth]{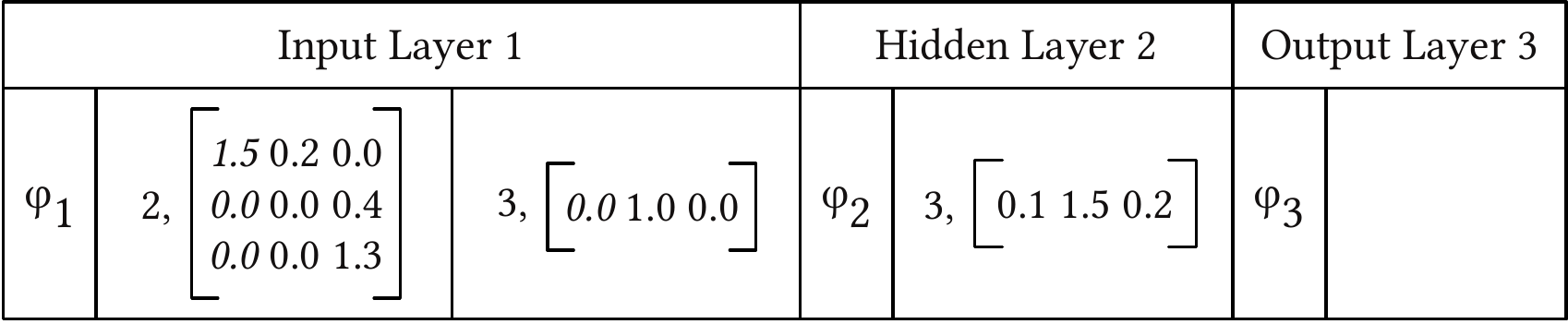}
    \label{fig:alf-geneticEncoding-genotype}}
  \hfil
  \subfloat[Phenotype]{\includegraphics[width=0.32\textwidth]{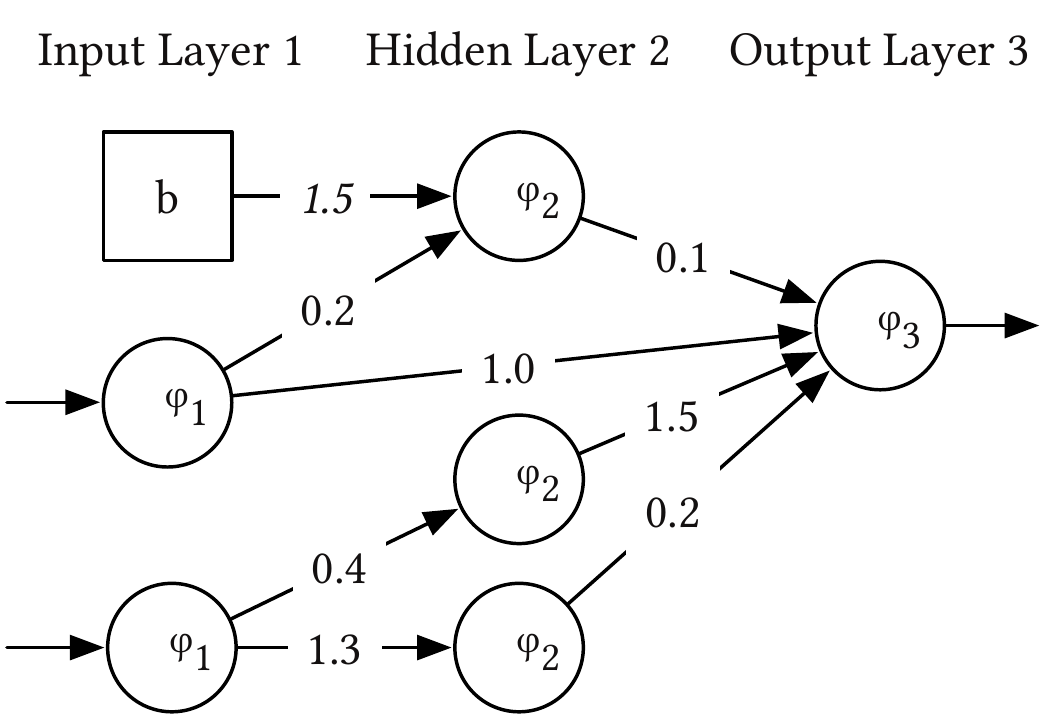}
    \label{fig:alf-geneticEncoding-phenotype}}
  \caption{Example of genotype-phenotype mapping in ALF}
  \label{fig:alf-geneticEncoding-genotypePhenotypeMapping}
\end{figure}

ALF has the ability to evolve topologies of different sizes. In this context, a zero entry in the connection matrix is interpreted as \emph{no~connection}. \Cref{fig:alf-geneticEncoding-genotypePhenotypeMapping} shows an example of a corresponding genotype-phenotype mapping. There is an input, a hidden, and an output layer as well as a bias unit~$b$. The input layer with the id 1 contains two nodes and a layer connection to the hidden layer as well as a bias connection encoded in the first column, and a layer connection to the output layer. All possible bias connections are highlighted in italics. The hidden layer with the id 2 contains three nodes and a layer connection to the output layer. The output layer contains one node. \Cref{fig:alf-geneticEncoding-genotype} shows the corresponding encoding including the activation functions of the layers. If the individual connections are considered, a zero connection can be detected between the input~($j = 2$) and the hidden node~($i = 2$), \ie, one that is not expressed in the phenotype as shown in~\Cref{fig:alf-geneticEncoding-phenotype}.
One of the main problems of NE is the permutations problem~\cite{hancock1997genetic}, \ie, there is more than one possibility to express a solution to a weight optimization problem with an ANN. Specifically, if genotypes represent the same solution but do not have the same encoding, the crossover is likely to reproduce offspring with defective genotypes. For example, crossing over two structurally identical but differently encoded ANNs could result in offspring in which critical information is lost or, respectively, genes are duplicated. To prevent this, there is at most one layer connection between one source and destination layer in ALF, whereby this layer connection is assigned to the source layer. The layer connections of a layer are sorted in ascending order by destination layer id. Due to the matrix data structure, at most one cell entry can exist as a connection between two nodes. Therefore, the genetic encoding in ALF is unique, allowing efficient comparisons between genes.

\subsection{Semantic and Structural Speciation~(SSS)}
\label{sec:alf-sss}

In TWEANN, it is unlikely that a new connection immediately expresses a useful function. Optimization often requires several generations, whereby the new structure does not have the necessary time to evolve. ANNs with such innovations need to be protected so that they have enough time to express these in useful ways. For this purpose, concepts of species were introduced in~\cite{holland1992adaptation, stanley2002evolving}.

A species is a group of individuals with similar characteristics that compete mainly within their own species. Thus, innovations are protected because they are not dominated by already optimized structures. To determine whether two individuals belong to the same species, a compatibility function~$\delta$ is needed. While in~\cite{stanley2002evolving} a topological comparison is made for their classification, we introduce the structural and semantic comparison in ALF.

\begin{figure}[t]
  \centering
  \subfloat[ANNs before alignment]{\includegraphics[width=0.23\textwidth]{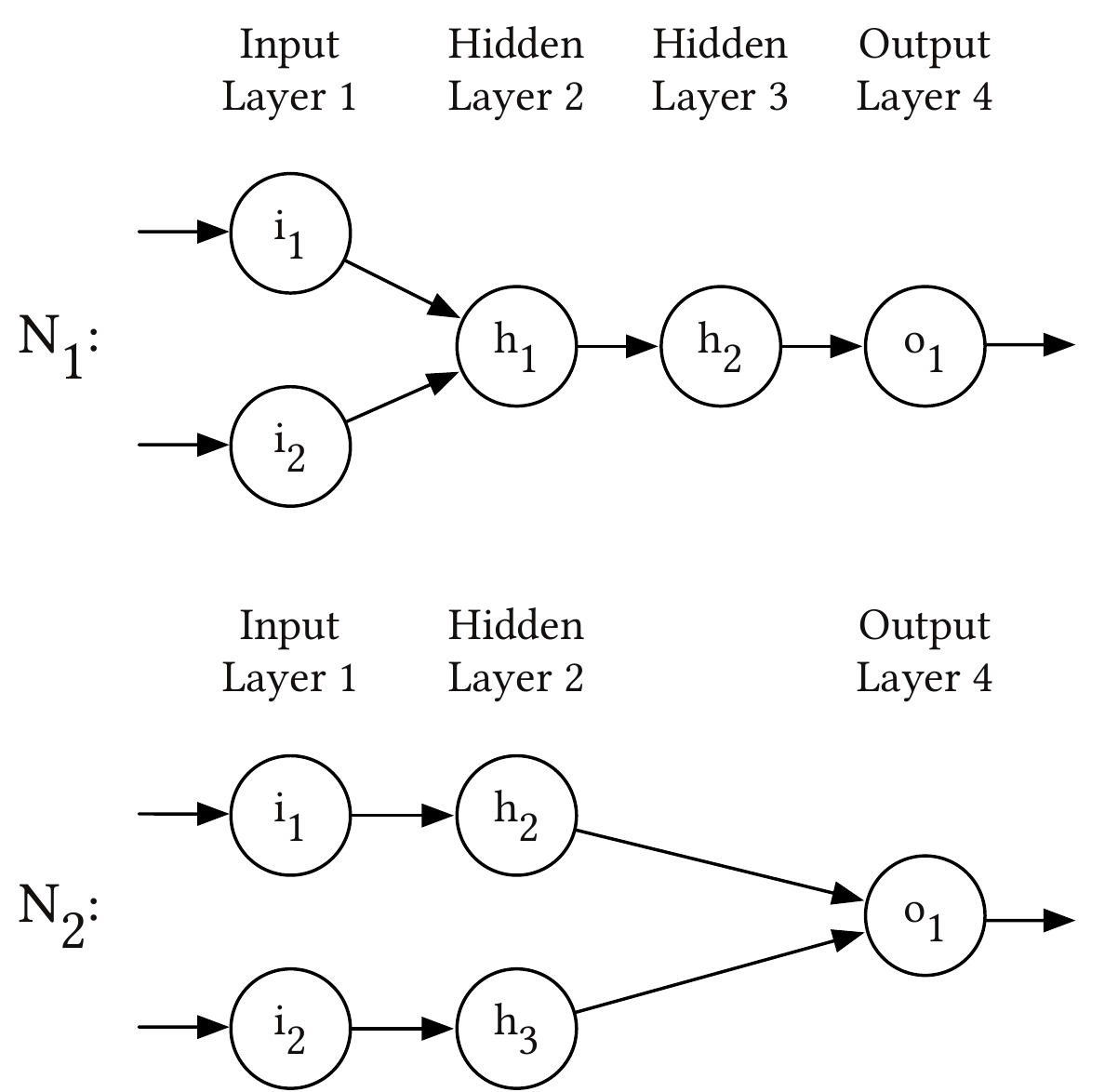}
    \label{fig:alf-sss-annsBeforeAlignment}}
  \hfil
  \subfloat[ANNs after alignment]{\includegraphics[width=0.23\textwidth]{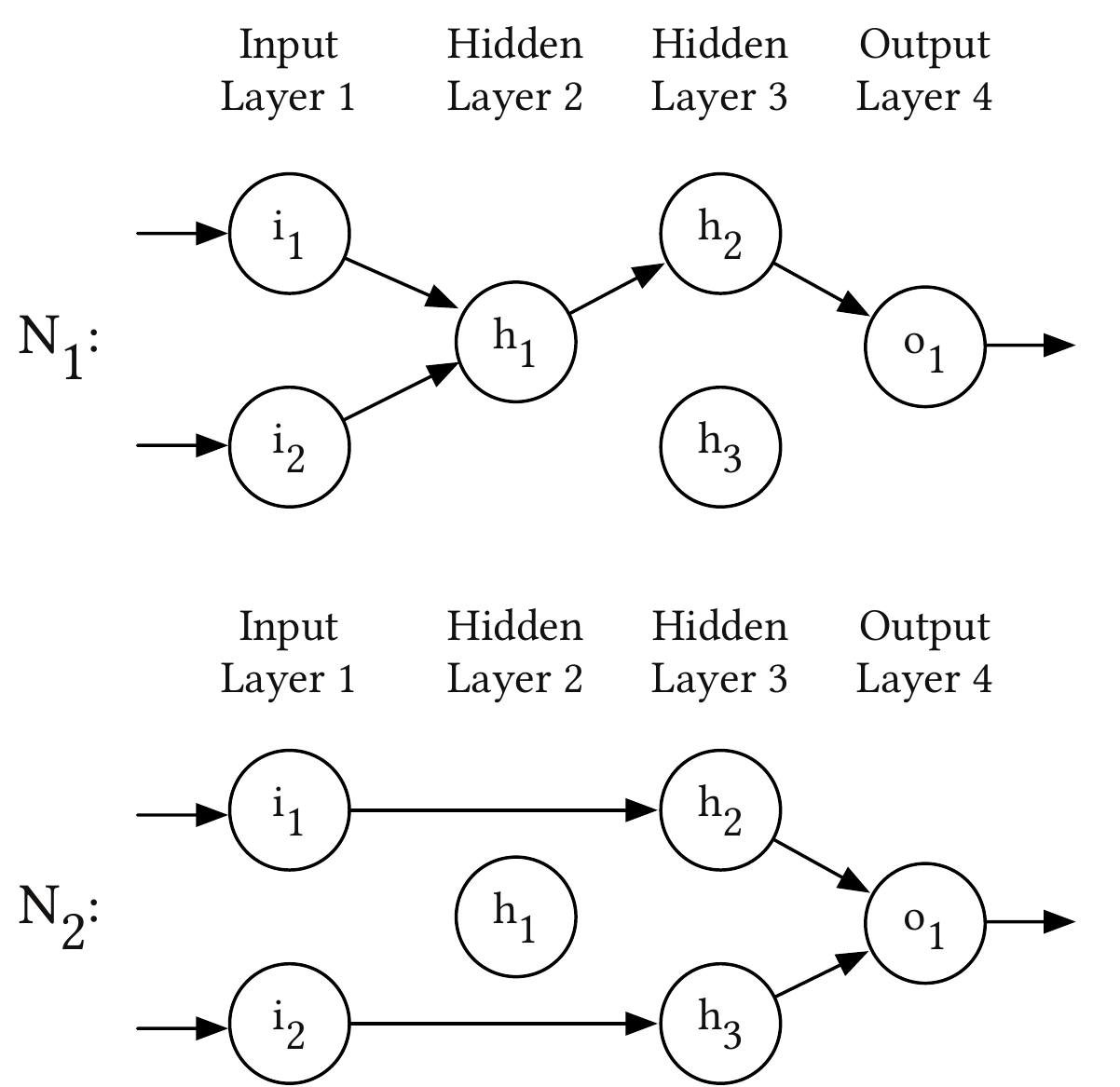}
    \label{fig:alf-sss-annsAfterAlignment}}
  \caption{Example of structural comparison of ANNs}
  \label{fig:alf-sss-structuralComparison}
\end{figure}

Because of the solution of the permutations problem in~\Cref{sec:alf-geneticEncoding}, the structural similarity~$T$ can be measured efficiently. This comparison is performed by aligning the layer structures via the union of nodes. This preserves the semantics of ANNs to be compared, as no weighted connections are added. Then, a layer-wise structural comparison between ANNs can be performed by counting the shared connections~$E_{\mathit{shared}}$ and maximum connections~$E_{\mathit{max}}$, implicitly including the nodes. \Cref{fig:alf-sss-structuralComparison} shows an example of the structural comparison. Specifically, two ANNs~$N_1, N_2$, which are to be compared, are shown in \Cref{fig:alf-sss-annsBeforeAlignment}, where $N_2$ is considered a smaller ANN because it has fewer layers compared to $N_1$. For simplicity, the nodes are labeled accordingly. As a result of the union of nodes, in addition to the input and output layers, the hidden layer 3 of $N_2$ is compared with the hidden layer 3 of $N_1$. Due to the unique encoding described in \Cref{sec:alf-geneticEncoding}, the nodes $h_2$ of $N_1$ and $N_2$ are structurally identical and are merged, which can be seen in \Cref{fig:alf-sss-annsAfterAlignment}. Obviously, there is only one shared connection from $h_2$ to $o_1$. The other connections are not shared. Thus, from $E_{\mathit{shared}} = 1$ and $E_{\mathit{max}} = 4$ it follows that the structural similarity is $T = \frac{E_{\mathit{shared}}}{E_{\mathit{max}}} = \frac{1}{4} = 0.25 \Rightarrow 25 $\,\%.
The semantic comparison~$B$ provides information about the similarity of the behavior of the considered ANNs. The behavior of an ANN is defined by its predictions on given inputs. The inputs can be data from a dataset or data collected at runtime. The similarity of the ANNs' predictions to be compared is determined by the correlation coefficient~$\rho$, as shown in \Cref{eq:alf-sss-correlationCoefficient}

\begin{equation}
  \label{eq:alf-sss-correlationCoefficient}
  \rho_{X, Y} = corr(X, Y) = \frac{\sum_{i=1}^n (X_i - \bar{X})(Y_i -
                             \bar{Y})}{\sqrt{\sum_{i=1}^n (X_i - 
                             \bar{X})^2} \cdot \sqrt{\sum_{i=1}^n (Y_i -
                             \bar{Y})^2}}
  \textrm{,}
\end{equation}

where $X$ refers to the predictions of one ANN and $Y$ refers to the predictions of another ANN. The correlation coefficient measures the linear dependence of the individuals. Since similarity is characterized only by correlation, anti-correlation is disregarded. For this reason, the function $B = max(0, \rho_{X, Y})$ applies. The significance of the correlation is tested using the t-statistic~\cite{schober2018correlation}. Since there is not necessarily normally distributed data, we apply the central limit theorem~\cite{anderson2010central} from a sample size of 30~samples. Furthermore, we require at least a 80\,\% confidence level for the correlation to be considered significant.

While structural species comparison protects topological diversity, semantic comparison protects diversity of behavior patterns. Both factors contribute to increasing genetic diversity. Depending on the problem, the importance of both factors can be adjusted by the coefficients $c_1$ and $c_2$, which is shown in \Cref{eq:alf-sss-compatibilityFunction}
\begin{equation}
  \label{eq:alf-sss-compatibilityFunction}
  \delta = \frac{T \cdot c_1 + B \cdot c_2}{c_1 + c_2}
  \textrm{.}
\end{equation}
The compatibility measure allows a simple speciation of the population. In each generation, the offspring are classified into species. Each species is represented by the containing fittest individual~(mascot) from the previous generation. An offspring is compared with each mascot and assigned to the most similar species. In this context, the compatibility threshold~$\delta_t$ indicates the minimum similarity required. If $\delta \geq \delta_t$ does not apply to any species, a new species is created and the offspring becomes the mascot.

The objective of the component SSS is accordingly supporting a diversity of complex structures and behavior patterns to form better individuals in order to perform the search for a solution efficiently.

\subsection{Dynamic Adaptation of Population~(DAP)}
\label{sec:alf-dap}

In some TWEANN algorithms, the initial population consists of randomly large topologies so that topological diversity exists immediately. However, this may result in individuals with already large ANN topologies whose fitness is higher than the fitness of individuals with smaller topologies. Such ANNs initially dominate the smaller ANNs and make the problem of finding minimal solutions more difficult, since the larger ANNs are more likely to evolve, although the smaller ANNs may also have the potential to solve the respective problem. In addition, ANNs which are topologically different but semantically identical can result. Thus, the ANN of each individual of the initial population in ALF is minimally initialized, \ie, consisting of an input and output layer with one layer connection of random weights. This reduces the probability of the dominance, but still provides sufficient initial genetic diversity.

Another problem can occur if a population size cannot change within the generations, as this may lead to stagnation of exploration in unsuitable search regions that do not advance learning. Therefore, a dynamic population is introduced in ALF, \ie, the population can increase and decrease, which is described as follows.

In ALF, fitness sharing is used for the reproduction mechanism, \ie, a species~$s_i$ contains the average fitness~$f_{s_i}$ of its assigned individuals, where $i = 1, \dots, n$ and $n$ is the number of species. The fitness proportion of the species is denoted as $F_{s_i} = \frac{f_{s_i}}{\sum_{i=1}^n f_{s_i}}$. In addition, each species is assigned an age~$A_{s_i}$, which identifies how many generations this very species has been in the population. The age of the oldest species is denoted as~$A_o$. To decide whether a species is to be deleted, both factors are considered in combination, which is shown by \Cref{eq:alf-dap-fitnessAgeCondition}
\begin{equation}
  \label{eq:alf-dap-fitnessAgeCondition}
	F_{s_i} < \frac{1}{n} \cdot c_3 \wedge A_{s_i} > A_o \cdot c_4
  \textrm{,}
\end{equation}
where $c_3$ and $c_4$ are coefficients for adjusting the importance of these factors. Age dependency in combination with the fitness proportion penalizes older species while protecting younger species. Older species, which have a lower proportion of the population's fitness compared to other species, are treated like so-called stale species. A species is called stale if it has not improved over a predetermined number \textit{staleness} of generations. These unsuccessful species are therefore deleted.

Furthermore, the weakest individuals of all species are deleted, whereby their number depends on $F_{s_i}$ of the respective species and does not exceed 50\,\% of the population. The population is then replenished by the offspring of the remaining individuals in each species. The new population size $P_{\mathit{new}}$ depends on the proportion of deleted species $S_{\mathit{del}}$, which is shown by \Cref{eq:alf-dap-newPopSize}
\begin{equation}
  \label{eq:alf-dap-newPopSize}
  P_{new} = \begin{cases}
              min(P_{\mathit{max}},\,P_{\mathit{old}} \cdot (1 +
                S_{\mathit{del}})), & S_{\mathit{del}} > 0\\
		          max(P_{\mathit{old}} - P_{\mathit{init}} \cdot \mathit{ebb},
                P_{\mathit{init}}), & S_{\mathit{del}} = 0
	          \end{cases}
  \textrm{,}
\end{equation}
where $\mathit{ebb}$ denotes an ebb factor. Thus, the greater the number of species deleted, the larger the population becomes, whereby the new population size~$P_{\mathit{new}}$ is limited by a predetermined value~$P_{\mathit{max}}$. Therefore, more room is available for species to reproduce, so that search regions can be sampled more frequently to increase the probability of leaving insufficient search regions more rapidly. The ebb factor ensures that the population size is reduced if no species are deleted, whereby the population never becomes smaller than the initial population~$P_{\mathit{init}}$. Since weaker individuals are deleted, fitter offspring resulting from reproduction still get sufficient time to solve the sub-problem. This also reduces unnecessary evaluations of weaker individuals.

The species also reproduce depending on their fitness. The number of respective offspring $O_{s_i}$ is determined by \Cref{eq:alf-dap-breed}
\begin{equation}
  \label{eq:alf-dap-breed}
  O_{s_i} = F_{s_i} \cdot (1 - o_{\mathit{init}}) \cdot (P_{\mathit{new}} -
    (P_{\mathit{max}} - P_{\mathit{vacant}}))
  \textrm{,}
\end{equation}
where $P_{\mathit{vacant}}$ is the available room that can be used for the reproduction of offspring and $o_{\mathit{init}}$ is the proportion of newly initialized ANNs. Within the respective species, the fittest individuals with a higher probability are then selected as parents to reproduce offspring. The remaining room is replenished with those newly initialized ANNs. Thus, even in the ongoing evolutionary process, new ANNs are provided so that possible dead ends can be overcome with a higher probability via new innovations.

For this reason, the objective of the component DAP is to improve convergence properties by overcoming insufficient search regions through increased simultaneous sampling.

\subsection{Fitness-Based Genetic Operators~(FBGO)}
\label{sec:alf-fbgo}

On the basis of the selected parents from the mating pool described in~\Cref{sec:alf-dap}, variation operators are applied to reproduce offspring and explore the search space. These include mutation and crossover. In a standard mutation, genes are changed randomly, which can have a negative impact on the exploration in the search space. While small changes can increase the time required to explore the search space, large changes are not suitable for local optimization and can, thus, increase the time required to optimize found solutions from already explored search regions. If crossover is used, swapping genes of unequally fit individuals may result in unsuitable genes being inherited, leading to slow convergence. To counteract these problems, ALF introduces fitness-based mutation and crossover, which are described in the following.

A mutation in ALF can change both connection weights and ANN structures. A fitness-based mutation rate is calculated for all mutations, which determines how often ALF attempts to perform a respective mutation. The mutation rate $m_r$ is defined as \Cref{eq:alf-fbgo-mutationRate}
\begin{equation}
  \label{eq:alf-fbgo-mutationRate}
  m_r = min\left(max\left(1, \left(\left(\left(1 + m_o\right) -
        \frac{F}{f_t}\right) \cdot m_a\right)\right), m_a\right)
  \textrm{,}
\end{equation}
where $m_o$ is a predefined mutation offset. On the one hand, in relation to the fitness threshold~$f_t$, $m_o$ indicates up to which proportion of the current fitness~$F$ of an individual the maximum number of attempts $m_a$ is performed. On the other hand, it ensures that $m_r$ cannot become 0, so that mutations still take place even with high fitness values. A connection weight~$a_{ij}$ is changed by adding a random number from the normal distribution~$\mathcal{N}\left(0, 1.25 - \frac{F}{f_t}\right)$. The fitter an individual is, the smaller changes are made, \ie, when $F$ is low, compared to $F_t$, the search process has more influence and when $F$ is high, the optimization process is strengthened. Based on the genetic encoding~(see~\Cref{sec:alf-geneticEncoding}), the \emph{weight mutation} creates, changes or deletes a connection. In order to change connections significantly more often than to add layers or nodes, it is determined with regard to the \emph{node} and \emph{layer mutation} in addition to $m_r$ whether such a respective mutation should be performed. The corresponding probability calculation~$\mathsf{P}_{\mathit{mut}}$ is defined in \Cref{eq:alf-fbgo-mutationProb}
\begin{equation}
  \label{eq:alf-fbgo-mutationProb}
  \mathsf{P}_{\mathit{mut}} = min\left(1 + m_o -
    \frac{\frac{E}{E_{\mathit{full}}} + \frac{F}{f_t}}{2}, 1\right)
  \textrm{,}
\end{equation}
where $E$ is the current number of connections and $E_{\mathit{full}}$ is the number of possible connections (assuming the ANN is fully connected). The proportion of connections to possible connections describes the potential for new connections, \ie, new connection possibilities~(nodes or layers) become more likely to be added if $\frac{E}{E_{\mathit{full}}}$ is closer to 1. \emph{Node mutation} randomly adds nodes into hidden layers, determining the number of nodes~$V_{\mathit{mut}}$ as shown in \Cref{eq:alf-fbgo-nodeMutation}
\begin{equation}
  \label{eq:alf-fbgo-nodeMutation}
  V_{\mathit{mut}} = \left\lceil V_{\mathit{max}} - \frac{F}{f_t} \cdot
    V_{\mathit{max}} \right\rceil
  \textrm{,}
\end{equation}
where $V_{\mathit{max}}$ is predetermined and is the maximum number of nodes to be added. The \emph{layer mutation} inserts a new layer connection or a new hidden layer. For this purpose, a layer is randomly selected and checked whether a layer connection can still be added. If not, a new hidden layer is appended. The number of nodes of the new layer is calculated as in \Cref{eq:alf-fbgo-nodeMutation}.

\begin{figure}[t]
  \centering
  \includegraphics[width=0.43\textwidth]{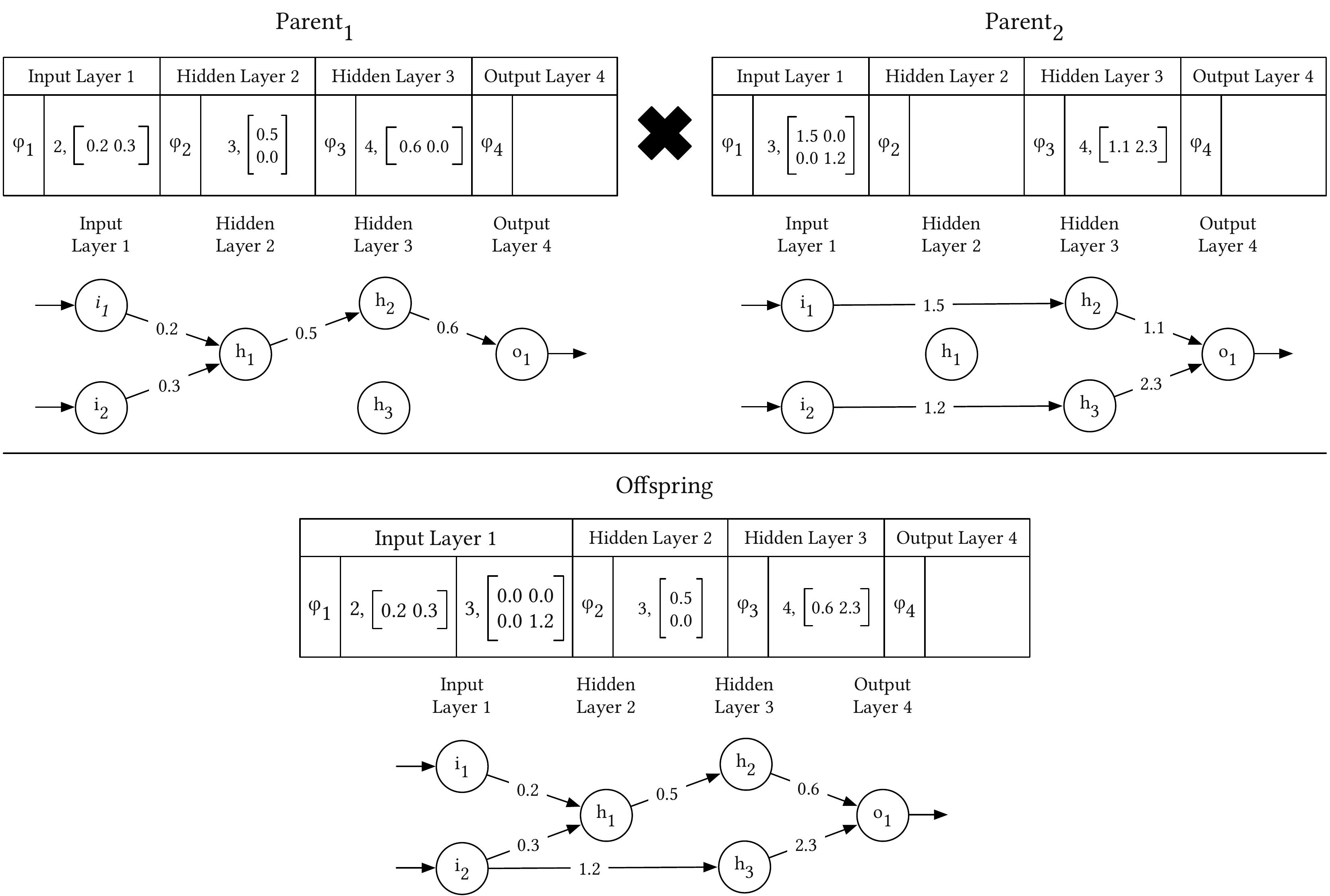}
  \caption{Example of crossing over genes of parents}
  \label{fig:alf-fbgo-crossover}
\end{figure}

Mutations will gradually increase the size of genotypes in ALF, resulting in genotypes of different sizes. In order to be able to compare such layer structures, ALF uses layer alignment by node union as described in~\Cref{sec:alf-sss}. Based on the solution of the permutations problem as described in~\Cref{sec:alf-geneticEncoding}, it is, thus, possible to identify gene matches efficiently. In ALF, there is a hyper-parameter~$\mathsf{P}_{\mathit{cross}}$ that indicates the probability of a crossover between selected individuals. When crossing over, fitness has an impact on the inheritance. Therefore, before crossing over, the respective fitness proportion is determined in relation to the fitness sum of both parents. The fitness proportion is interpreted as a probability during the gene comparison. Accordingly, when matching genes are compared, the gene of the fitter parent is more likely to be inherited. When disjoint genes are considered, a gene of the fitter parent is more likely to be inherited, while a gene of the weaker parent is more likely to be not inherited. Consider \Cref{fig:alf-fbgo-crossover}, which shows an example of the crossover used in ALF assuming that parent $\mathit{Parent}_1$ is fitter than $\mathit{Parent_2}$. While all disjoint genes are inherited from $\mathit{Parent_1}$, the gene from node $i_1$ to $h_2$ is not inherited from $\mathit{Parent_2}$. For the matching gene from $h_2$ to $o_1$, the higher fitness of $\mathit{Parent}_1$ causes its gene to be inherited.
By fitness-based adding of new genes to the population and crossing over genotypes, the objective of the component FBGO is to increase the probability of optimization success of individuals.

\section{Experimental Results}
\label{sec:experimentalResults}

In this section, ALF is analyzed empirically. For this purpose, we implemented the components described in the previous section in~C++17 and measured their performances in terms of the average time in 
minutes~(t in min), average number of generations~(G), connections~(E), and hidden nodes~(H) required by applying them to large-scale RL problems. The analysis of ALF aims at answering the following questions:
\begin{inparaenum}
  \item Does the symbiosis of the proposed components of ALF lead to an
        effective evolution of necessary ANNs?
  \item Does ALF find solutions efficiently?
\end{inparaenum}
The first question establishes that the symbiosis of ALF's components indeed reduces the memory usage and runtime. To answer the second question, in this paper we choose the TWEANN algorithm \emph{NeuroEvolution of Augmenting Topologies}~(NEAT)~\cite{stanley2002evolving} as test algorithm for the evaluation, because it has already proven to be successful in complex control learning tasks~\cite{stanley2003competitive} and RAM-based Atari games, and was more reliable than other TWEANN algorithms like HyperNEAT~\cite{hausknecht2014neuroevolution}. Compared to other NE algorithms, NEAT is characterised in particular by topology evolution and the protection of innovations via speciation by topological comparisons of genotypes. We limit ourselves to NE because the focus is on achieving better performance of evolving ANNs and NE algorithms demonstrated similar performance in games compared to RL methods such as deep Q-learning in the literature, whereby RL methods required significantly more training time~\cite{such2017deep}.

\subsection{SMB as a Benchmark Task}
\label{sec:experimentalResults-smb}

\emph{Super Mario Bros.}~(SMB) is a game for the \emph{Nintendo Entertainment System}~(NES) that is divided into several game worlds with different levels~(problems). In each level, the protagonist~(Mario) starts from a specific location and his objective is to reach the flagpole or axe in a certain time, which leads to the next level. Occasionally, a level contains hidden warp pipes, which let him skip some levels. In addition, there are various challenges such as canyons, enemies or dead ends that Mario has to overcome. Another game element is the loop command object, an invisible tile that resets Mario's position by several screens if he does not follow a right path. If Mario dies by walking into an enemy or falling into a canyon, he loses one life and has to start from the beginning of the level. If he reaches a checkpoint, it becomes the new starting point of the level.

The criterion for success on this task was the completion of the levels considered. The task was suitable to test ALF's abilities because of several reasons. SMB is a sequential and unlabeled problem, which is not linearly separable. Thus, hidden nodes have to be evolved. Furthermore, the task of reaching the flagpole or axe is PSPACE-complete~\cite{demaine2016super}, whereby a large solution space with few suitable solutions exists unlike in other RL problems like cart-pole~\cite{nagendra2017comparison}.

To consider many game elements, three game worlds were studied and considered as large-scale RL problems: 4-2, 4-4, and 7-2. World 4-2 contains canyons, enemies and passages, in which Mario has to wait a certain time. Moreover, this level contains a possible dead end, where Mario has to go back and jump at the correct time to overcome a high obstacle. World 4-4 contains the mentioned loop command objects. For both levels, a fitness function was defined as in \Cref{eq:experimentalResults-smb-fitnessHorizontal}

\begin{equation}
  \label{eq:experimentalResults-smb-fitnessHorizontal}
  f(d, q, x) = d + q + x
  \textrm{,}
\end{equation}

where $d$ is the game progress, $q$ is the scroll amount, and $x$ denotes the $x$-coordinate of Mario's position. While the fitness threshold \mbox{$f_t =$ 3,163} of world 4-2 referred to the $x$-coordinate of the flagpole, \mbox{$f_t =$ 2,770} of world 4-4 corresponded to the $x$-coordinate of the axe.

World 7-2 is a water level, where Mario can also move vertically. Therefore, the $y$-coordinate of Mario's position was additionally set in relation to his $x$-coordinate, which was defined by \Cref{eq:experimentalResults-smb-fitnessVertical}
\begin{equation}
  \label{eq:experimentalResults-smb-fitnessVertical}
  f(d, q, x, y) = d + q + 255x + (255 - y)
  \textrm{.}
\end{equation}
The fitness threshold of world 7-2 corresponded to the \\$x$-coordinate of the flag pole, \ie, \mbox{$f_t =$ 3,404}.

Regarding the activation function, we oriented ourselves on the sigmoid variant of NEAT~\cite{stanley2002evolving} and modified it to support inhibitory signals. The function is shown in \Cref{eq:experimentalResults-smb-modSigmoid}. It squashes the incoming values $v$ into the range $[-1.0, 1.0]$. To reduce the dimension of the learning problem, a bias unit was configured.
\begin{equation}
  \label{eq:experimentalResults-smb-modSigmoid}
  \phi(v) = sig_m(v) = \frac{2}{1 + e^{-4.9v}} - 1
\end{equation}
The NES has a frame consisting of 184,320~pixels. To handle this large input space, we compressed it into 80 inputs, which we considered as sufficiently small for our empirical study. We read out a field of view around Mario's position from the memory, whereby each enemy and game element was encoded differently to distinguish them. A frame is divided into blocks and tiles. A tile consists of $8 \times 8$~pixels and a block consists of $2 \times 2$ tiles. We chose a field of view of $10 \times 8$~blocks, \ie, we chose an area of $x = [-2, 8]$ and $y= [-5, 3]$ blocks relative to Mario's position. This number of blocks was determined through experiments and observations of the level structures of SMB. Each block represented a feature, which was determined by the mean of the tile values within the block. Enemies were encoded as negative values, while other game elements were encoded as positive values. Mario was encoded as 0. Each encoded value was scaled to a range of $[-1.0, 1.0]$ before being passed to the ANN in order to reduce feature dominance and, therefore, possible generalization errors. Moreover, the number of output nodes was set to 6, whereby each node represented a button on the NES controller. These were the buttons that the ANN had to press or release after each processed frame.

All experiments were conducted on an Intel Xeon E3-1230 v6 with 3.2~GHz and 16~GB of main memory running Fedora 22. Because of the complexity of SMB, 10 runs were performed and the respective average, possible extremities, and divergences were determined. The timeout to finish a level was set to 2~days.

\subsection{Verification of ALF's Components}
\label{sec:experimentalResults-verification}

In this section, it is verified whether the introduced components of ALF are reliable and their symbiosis leads to reduced runtime and memory usage. Therefore, ALF was applied to the SMB world 4-2.

\subsubsection{Hyper-parameter Settings}
\label{sec:experimentalResults-verification-hyperparameterSettings}

For all experiments, the same experimental settings were used. However, the parameters were not adapted to SMB. \Cref{tab:experimentalResults-verification-hyperparameterSettings-alf} shows the most important hyper-parameters used in ALF, which were determined experimentally and are described in the following.
\begin{table}[t]
  \centering
  \caption{Hyper-parameter settings for ALF}
  \label{tab:experimentalResults-verification-hyperparameterSettings-alf}
  \begin{tabularx}{\columnwidth}{Xr}
    \toprule
    \textbf{Hyper-parameter} & \textbf{ALF} \\
    \toprule
    Initial population size~($P_{init}$) & 150 \\
    Maximum population size~($P_{max}$) & 300 \\
    Ebb factor~($ebb$) & 0.05 \\
    Mutation offset~($m_o$) & 0.1 \\
    Maximum mutation attempts~($m_a$) & 5 \\
    Maximum added nodes~($V_{max}$) & 5 \\
    Crossover probability~($\mathsf{P}_{\mathit{cross}}$) & 0.7 \\
    Staleness~(\textit{staleness}) & 15 \\
    Oldest species age coefficient~($c_4$) & 0.5 \\
    Species fitness proportion coefficient~($c_3$) & 0.4 \\
    Newly initialised ANNs proportion~($o_{init}$) & 0.1 \\
    Semantic coefficient~($c_2$) & 0.75 \\
    Structural coefficient~($c_1$) & 0.25 \\
    Compatibility threshold~($\delta_t$) & 0.3 \\
    \bottomrule
  \end{tabularx}
\end{table}
The initial population size consists of 150~individuals, with a maximum size of 300~individuals. This provides sufficient room for more species, allowing for greater genetic diversity. If no species is deleted, the population shrinks by 5\,\%, whereby the population size never gets smaller than the initial population size. Otherwise, the population grows in relation to the number of deleted species to provide additional individuals. This ensures that the exploration of the search space is never reduced if there is an approximation to an insufficient local optimum. 

Until a fitness proportion of 10\,\% in relation to the fitness threshold is reached, the maximum number of mutation attempts is performed on the genes. Otherwise, the number of attempts is decreased in relation to the fitness and number of connections, in order to approximate to a local optimum more purposefully. Since connections are to be added more frequently than nodes, a proportionally low maximum number of nodes is chosen with respect to the mutation. The probability of reproducing an offspring without any crossover is 30\,\%, which is common for GAs. 

If the maximum fitness of a species does not improve within 15~generations, the species is considered as stale and is deleted. The species is also deleted if it is older than half of the oldest species and its fitness proportion is smaller than 40\,\%, assuming the fitness is equally distributed. Thus, a species has enough time to evolve, but is penalized if it does not. One exception is a species that contains the best individual to prevent degradation of the best fitness in the population. The weakest individuals of each species are deleted, whereby the number of deleted individuals depends on the adapted species' fitness. However, the number of deleted individuals never exceeds 50\,\% of the population size and the best individual of a species is never deleted. Each new population is replenished with an offspring proportion of 90\,\%. The behavior of the ANNs is considered more important than their structural information. Therefore, the semantic coefficient is set to 75\,\%, while the structural coefficient is set to 25\,\%. In this context, it is required that an offspring, compared to a mascot, has to achieve a minimum similarity of 30\,\% to be assigned. Otherwise, a new species is created.

\subsubsection{Results}
\label{eq:experimentalResults-verification-results}

Based on the initial run~(\emph{no component}) without using any of the addressed components
\begin{inparaenum}
  \item SSS,
  \item DAP, and
  \item FBGO,
\end{inparaenum}
we aimed at determining whether the SMB world 4-2 can be solved faster with fewer average connections and hidden nodes in fewer generations. The results are summarized in~\Cref{tab:experimentalResults-verification-results-symbioses} and described in the following.\footnote{Note when SSS was deactivated, each individual was assigned to a separate species.}
\begin{table}[t]
  \centering
  \fontsize{6.8}{7.9}\selectfont
  \caption{Comparison between component symbioses of ALF in terms of runtime and ANN sizes to solve SMB world 4-2}
  \label{tab:experimentalResults-verification-results-symbioses}
  \begin{tabular}{lcccc}
    \toprule
    ~ & \multicolumn{4}{c}{SMB}\\
    \cmidrule(lr){2-5}
    Symbiosis of ALF's components & \multicolumn{4}{c}{World 4-2}\\
    \cmidrule(lr){2-5}
    ~ & t in min & G & E & H\\
    \midrule
    No component & 1,801 & 603 & 4,288 & 1,316\\
    SSS & 1,363 & 334 & 3,522 & 1,021\\
    DAP & 1,789 & 572 & 4,109 & 1,193\\
    FBGO & 1,456 & 478 & 3,768 & 687\\
    SSS+DAP & 1,073 & 241 & 3,004 & 980\\
    SSS+FBGO & 1,106 & 256 & 2,703 & 389\\
    DAP+FBGO & 1,194 & 328 & 3,235 & 546\\
    SSS+DAP+FBGO & 978 & 194 & 2,445 & 304\\
    \midrule[\heavyrulewidth]
    \multicolumn{1}{l}{\footnotesize t in min} & \multicolumn{4}{l}{\footnotesize Time in minutes}\\
    \multicolumn{1}{l}{\footnotesize G} & \multicolumn{4}{l}{\footnotesize Number of generations needed}\\
    \multicolumn{1}{l}{\footnotesize E} & \multicolumn{4}{l}{\footnotesize Number of evolved connections of the ANNs}\\
    \multicolumn{1}{l}{\footnotesize H} & \multicolumn{4}{l}{\footnotesize Number of evolved hidden nodes of the ANNs}\\
    \multicolumn{1}{l}{\footnotesize SSS} & \multicolumn{4}{l}{\footnotesize Semantic and Structural Speciation}\\
    \multicolumn{1}{l}{\footnotesize DAP} & \multicolumn{4}{l}{\footnotesize Dynamic Adaptation of Population}\\
    \multicolumn{1}{l}{\footnotesize FBGO} & \multicolumn{4}{l}{\footnotesize Fitness-Based Genetic Operators}\\
    \bottomrule
	\end{tabular}
\end{table}
ALF is consistent in finding a solution. In particular, the symbiosis of the components results in less time, generations, connections, and hidden nodes required on average to solve this problem.

On average, the worst case without using any component needs 1,801~minutes, 603~generations, 4,288~connections, and 1,316~hidden nodes, while the best symbiosis~(SSS+DAP+FBGO) requires 978~minutes, 194~generations, 2,445~connections, and 304~hidden nodes. This reduces the ANN size by about 50.95\,\%. Furthermore, the problem is solved almost 2~times faster. In particular, the addition of FBGO significantly reduces the number of hidden nodes or, respectively the proportion between connections and hidden nodes. Therefore, \eg, the number of hidden nodes in SSS+FBGO is reduced by about 31.94\,\% and the number of hidden nodes in DAP+FBGO is reduced by about 28.69\,\%.

In the context of SSS, we observed that a species, which went through the bonus area, was evolved. Individuals of this species went through a bonus pipe and were able to solve the problem faster. With respect to DAP, we observed that population size increased when the population had to solve complex sub-problems like crossing canyons. Due to more individuals, these sub-problems could be solved faster, especially in symbioses with FBGO or SSS.

In conclusion, ALF solves SMB world 4-2 without any difficulties. Specifically, by adding the mentioned components (in symbioses), this problem is solved faster with less memory usage.

\subsection{Evaluation of ALF's Performance}
\label{sec:experimentalResults-evaluation}

To obtain meaningful performance comparisons with NEAT and to determine whether ALF is more efficient in finding solutions, SMB worlds 4-2, 4-4 and 7-4 are considered. First, the experimental setup is described, followed by the analysis of ALF's performance.

\subsubsection{Hyper-parameter Settings}
\label{sec:experimentalResults-evaluation-hyperparameterSettings}

For the following experiments we used the same hyper-parameter settings for ALF as described in~\Cref{sec:experimentalResults-verification-hyperparameterSettings}. Moreover, we set up NEAT as described in~\cite{hausknecht2014neuroevolution} and adjusted the hyper-parameters, \ie, the same available genetic parameters, such as population size, were matched accordingly to allow a representative comparison.

\subsubsection{Results}
\label{sec:experimentalResults-evaluation-results}
\begin{table}[t]
  \centering
  \fontsize{7.1}{8.1}\selectfont
  \caption{Comparison between NEAT and ALF in terms of runtime and ANN sizes to solve different problems}
  \label{tab:experimentalResults-evaluation-results-comparison}
  \begin{tabular}{ccccccccc}
    \toprule
    ~ & \multicolumn{8}{c}{Algorithm} \\
    \cmidrule(lr){2-9}
    SMB & \multicolumn{4}{c}{NEAT} & \multicolumn{4}{c}{ALF} \\
    \cmidrule(lr){2-5}
    \cmidrule(lr){6-9}
    ~ & t in min & G & E & H & t in min & G & E & H \\
    \midrule
    World 4-2 & 1,438 & 387 & 3,873 & 1,159 & 978 & 194 & 2,445 & 304 \\
    World 4-4 & 1,373 & 372 & 3,621 & 1,106 & 889 & 157 & 2,219 & 339 \\
    World 7-2 & 1,894 & 486 & 4,305 & 1,438 & 1,454 & 273 & 3,252 & 483 \\
    \midrule[\heavyrulewidth]
    \multicolumn{1}{r}{\footnotesize t in min} & \multicolumn{8}{l}{\footnotesize Time in minutes} \\
    \multicolumn{1}{r}{\footnotesize G} & \multicolumn{8}{l}{\footnotesize Number of generations needed} \\
    \multicolumn{1}{r}{\footnotesize E} & \multicolumn{8}{l}{\footnotesize Number of evolved connections of the ANNs} \\
    \multicolumn{1}{r}{\footnotesize H} & \multicolumn{8}{l}{\footnotesize Number of evolved hidden nodes of the ANNs} \\
    \bottomrule
  \end{tabular}
\end{table}
Next, we compare ALF with NEAT in order to determine whether ALF can find solutions efficiently. The results are summarized in \Cref{tab:experimentalResults-evaluation-results-comparison} and clearly confirm that the components proposed satisfy the objectives of this work. First, the proposed algorithm is capable of effectively solving the large-scale RL problems mentioned. Second, it addresses one of the known drawbacks of existing solutions: the evolution of inefficient ANNs for large-scale RL problems.

In this context, ALF outperforms NEAT in terms of runtime, generations, connections and hidden nodes required on each game world. In total, ALF solves the problems about 461~minutes faster on average and requires 207~generations less compared to NEAT. In addition, the ANN sizes are reduced by about 41.67\,\% on average.

Overall, the proposed algorithm is able to solve large-scale RL problems effectively, leading to efficient ANNs.

\section{Conclusion}
\label{sec:conclusion}

This work focused on one of the main challenges in NE -- the efficient solving of large-scale RL problems. The TWEANN algorithm ALF presented here can handle large-scale RL problems via the synergy of different components -- structural and semantic speciation, dynamic adaptation of observed candidate solutions, and integration of solution quality into genetic reproduction. Experimental results for large-scale RL problems confirmed that the proposed components are able to evolve efficient ANNs effectively. Future work will include optimizing the components described and extending the performance comparisons with other test algorithms and benchmark tasks.


\bibliography{alf}


\begin{thebibliography}{32}


\ifx \showCODEN    \undefined \def \showCODEN     #1{\unskip}     \fi
\ifx \showDOI      \undefined \def \showDOI       #1{#1}\fi
\ifx \showISBNx    \undefined \def \showISBNx     #1{\unskip}     \fi
\ifx \showISBNxiii \undefined \def \showISBNxiii  #1{\unskip}     \fi
\ifx \showISSN     \undefined \def \showISSN      #1{\unskip}     \fi
\ifx \showLCCN     \undefined \def \showLCCN      #1{\unskip}     \fi
\ifx \shownote     \undefined \def \shownote      #1{#1}          \fi
\ifx \showarticletitle \undefined \def \showarticletitle #1{#1}   \fi
\ifx \showURL      \undefined \def \showURL       {\relax}        \fi
\providecommand\bibfield[2]{#2}
\providecommand\bibinfo[2]{#2}
\providecommand\natexlab[1]{#1}
\providecommand\showeprint[2][]{arXiv:#2}

\bibitem[\protect\citeauthoryear{Anderson}{Anderson}{2010}]%
        {anderson2010central}
\bibfield{author}{\bibinfo{person}{Carolyn Anderson}.}
  \bibinfo{year}{2010}\natexlab{}.
\newblock \bibinfo{booktitle}{{\em Central Limit Theorem}}.
\newblock
\showDOI{%
\url{https://doi.org/10.1002/9780470479216.corpsy0160}}


\bibitem[\protect\citeauthoryear{Arulkumaran, Deisenroth, Brundage, and
  Bharath}{Arulkumaran et~al\mbox{.}}{2017}]%
        {arulkumaran2017brief}
\bibfield{author}{\bibinfo{person}{Kai Arulkumaran}, \bibinfo{person}{Marc
  Deisenroth}, \bibinfo{person}{Miles Brundage}, {and} \bibinfo{person}{Anil
  Bharath}.} \bibinfo{year}{2017}\natexlab{}.
\newblock \showarticletitle{A Brief Survey of Deep Reinforcement Learning}.
\newblock \bibinfo{journal}{{\em IEEE Signal Processing Magazine\/}}
  \bibinfo{volume}{34} (\bibinfo{date}{08} \bibinfo{year}{2017}).
\newblock
\showDOI{%
\url{https://doi.org/10.1109/MSP.2017.2743240}}


\bibitem[\protect\citeauthoryear{Berg and Whiteson}{Berg and Whiteson}{2013}]%
        {berg2013critical}
\bibfield{author}{\bibinfo{person}{Thomas Berg} {and} \bibinfo{person}{Shimon
  Whiteson}.} \bibinfo{year}{2013}\natexlab{}.
\newblock \showarticletitle{Critical factors in the performance of HyperNEAT}.
\newblock \bibinfo{journal}{{\em GECCO 2013 - Proceedings of the 2013 Genetic
  and Evolutionary Computation Conference\/}}, \bibinfo{pages}{759--766}.
\newblock
\showDOI{%
\url{https://doi.org/10.1145/2463372.2463460}}


\bibitem[\protect\citeauthoryear{Demaine, Viglietta, and Williams}{Demaine
  et~al\mbox{.}}{2016}]%
        {demaine2016super}
\bibfield{author}{\bibinfo{person}{Erik Demaine}, \bibinfo{person}{Giovanni
  Viglietta}, {and} \bibinfo{person}{Aaron Williams}.}
  \bibinfo{year}{2016}\natexlab{}.
\newblock \showarticletitle{Super Mario Bros. Is Harder/Easier than We
  Thought}.
\newblock  (\bibinfo{date}{06} \bibinfo{year}{2016}).
\newblock


\bibitem[\protect\citeauthoryear{Du and Swamy}{Du and Swamy}{2014}]%
        {du2014recurrent}
\bibfield{author}{\bibinfo{person}{K.-L Du} {and} \bibinfo{person}{M.N.s
  Swamy}.} \bibinfo{year}{2014}\natexlab{}.
\newblock \bibinfo{booktitle}{{\em Recurrent Neural Networks}}.
\newblock \bibinfo{pages}{337--353}.
\newblock
\showISBNx{978-1-4471-5570-6}
\showDOI{%
\url{https://doi.org/10.1007/978-1-4471-5571-3_11}}


\bibitem[\protect\citeauthoryear{Graves and Jaitly}{Graves and Jaitly}{2014}]%
        {graves2014online}
\bibfield{author}{\bibinfo{person}{A. Graves} {and} \bibinfo{person}{N.
  Jaitly}.} \bibinfo{year}{2014}\natexlab{}.
\newblock \showarticletitle{Towards end-to-end speech recognition with
  recurrent neural networks}.
\newblock \bibinfo{journal}{{\em 31st International Conference on Machine
  Learning, ICML 2014\/}}  \bibinfo{volume}{5} (\bibinfo{date}{01}
  \bibinfo{year}{2014}), \bibinfo{pages}{1764--1772}.
\newblock


\bibitem[\protect\citeauthoryear{Grossi and Buscema}{Grossi and
  Buscema}{2008}]%
        {grossi2008introduction}
\bibfield{author}{\bibinfo{person}{Enzo Grossi} {and} \bibinfo{person}{Massimo
  Buscema}.} \bibinfo{year}{2008}\natexlab{}.
\newblock \showarticletitle{Introduction to artificial neural networks}.
\newblock \bibinfo{journal}{{\em European journal of gastroenterology \&
  hepatology\/}}  \bibinfo{volume}{19} (\bibinfo{date}{01}
  \bibinfo{year}{2008}), \bibinfo{pages}{1046--54}.
\newblock
\showDOI{%
\url{https://doi.org/10.1097/MEG.0b013e3282f198a0}}


\bibitem[\protect\citeauthoryear{Hancock}{Hancock}{1997}]%
        {hancock1997genetic}
\bibfield{author}{\bibinfo{person}{Peter Hancock}.}
  \bibinfo{year}{1997}\natexlab{}.
\newblock \showarticletitle{Genetic Algorithms and permutation problems: a
  comparison of recombination operators for neural net structure
  specification}.
\newblock  (\bibinfo{date}{09} \bibinfo{year}{1997}).
\newblock


\bibitem[\protect\citeauthoryear{Hausknecht, Lehman, Miikkulainen, and
  Stone}{Hausknecht et~al\mbox{.}}{2014}]%
        {hausknecht2014neuroevolution}
\bibfield{author}{\bibinfo{person}{Matthew Hausknecht}, \bibinfo{person}{Joel
  Lehman}, \bibinfo{person}{Risto Miikkulainen}, {and} \bibinfo{person}{Peter
  Stone}.} \bibinfo{year}{2014}\natexlab{}.
\newblock \showarticletitle{A Neuroevolution Approach to General Atari Game
  Playing}.
\newblock \bibinfo{journal}{{\em Computational Intelligence and AI in Games,
  IEEE Transactions on\/}}  \bibinfo{volume}{6} (\bibinfo{date}{12}
  \bibinfo{year}{2014}), \bibinfo{pages}{355--366}.
\newblock
\showDOI{%
\url{https://doi.org/10.1109/TCIAIG.2013.2294713}}


\bibitem[\protect\citeauthoryear{Hochreiter}{Hochreiter}{1998}]%
        {hochreiter1998vanishing}
\bibfield{author}{\bibinfo{person}{Sepp Hochreiter}.}
  \bibinfo{year}{1998}\natexlab{}.
\newblock \showarticletitle{The Vanishing Gradient Problem During Learning
  Recurrent Neural Nets and Problem Solutions}.
\newblock \bibinfo{journal}{{\em International Journal of Uncertainty,
  Fuzziness and Knowledge-Based Systems\/}}  \bibinfo{volume}{6}
  (\bibinfo{date}{04} \bibinfo{year}{1998}), \bibinfo{pages}{107--116}.
\newblock
\showDOI{%
\url{https://doi.org/10.1142/S0218488598000094}}


\bibitem[\protect\citeauthoryear{Holland}{Holland}{1992}]%
        {holland1992adaptation}
\bibfield{author}{\bibinfo{person}{John~H. Holland}.}
  \bibinfo{year}{1992}\natexlab{}.
\newblock \bibinfo{booktitle}{{\em Adaptation in Natural and Artificial
  Systems: An Introductory Analysis with Applications to Biology, Control and
  Artificial Intelligence}}.
\newblock \bibinfo{publisher}{MIT Press}, \bibinfo{address}{Cambridge, MA,
  USA}.
\newblock
\showISBNx{0262082136}


\bibitem[\protect\citeauthoryear{Informatik, Bengio, Frasconi, and
  Schmidhuber}{Informatik et~al\mbox{.}}{2003}]%
        {informatik2003gradient}
\bibfield{author}{\bibinfo{person}{Fakultit Informatik}, \bibinfo{person}{Y.
  Bengio}, \bibinfo{person}{Paolo Frasconi}, {and} \bibinfo{person}{Jfirgen
  Schmidhuber}.} \bibinfo{year}{2003}\natexlab{}.
\newblock \showarticletitle{Gradient Flow in Recurrent Nets: the Difficulty of
  Learning Long-Term Dependencies}.
\newblock \bibinfo{journal}{{\em A Field Guide to Dynamical Recurrent Neural
  Networks\/}} (\bibinfo{date}{03} \bibinfo{year}{2003}).
\newblock


\bibitem[\protect\citeauthoryear{Kober, Bagnell, and Peters}{Kober
  et~al\mbox{.}}{2013}]%
        {kober2013reinforcement}
\bibfield{author}{\bibinfo{person}{Jens Kober}, \bibinfo{person}{J. Bagnell},
  {and} \bibinfo{person}{Jan Peters}.} \bibinfo{year}{2013}\natexlab{}.
\newblock \showarticletitle{Reinforcement Learning in Robotics: A Survey}.
\newblock \bibinfo{journal}{{\em The International Journal of Robotics
  Research\/}}  \bibinfo{volume}{32} (\bibinfo{date}{09} \bibinfo{year}{2013}),
  \bibinfo{pages}{1238--1274}.
\newblock
\showISBNx{978-3-642-27644-6}
\showDOI{%
\url{https://doi.org/10.1177/0278364913495721}}


\bibitem[\protect\citeauthoryear{Lavine and Blank}{Lavine and Blank}{2010}]%
        {lavine2010feed}
\bibfield{author}{\bibinfo{person}{B.K. Lavine} {and} \bibinfo{person}{T.R.
  Blank}.} \bibinfo{year}{2010}\natexlab{}.
\newblock \showarticletitle{Feed-Forward Neural Networks}.
\newblock \bibinfo{journal}{{\em Comprehensive Chemometrics\/}}
  \bibinfo{volume}{3} (\bibinfo{date}{01} \bibinfo{year}{2010}),
  \bibinfo{pages}{571--586}.
\newblock
\showISBNx{9780444527011}
\showDOI{%
\url{https://doi.org/10.1016/B978-044452701-1.00026-0}}


\bibitem[\protect\citeauthoryear{Lewicki and Marino}{Lewicki and
  Marino}{2004}]%
        {lewicki2004approximation}
\bibfield{author}{\bibinfo{person}{Grzegorz Lewicki} {and}
  \bibinfo{person}{Giuseppe Marino}.} \bibinfo{year}{2004}\natexlab{}.
\newblock \showarticletitle{Approximation by Superpositions of a Sigmoidal
  Function}.
\newblock \bibinfo{journal}{{\em Appl. Math. Lett.\/}}  \bibinfo{volume}{17}
  (\bibinfo{date}{12} \bibinfo{year}{2004}), \bibinfo{pages}{1147--1152}.
\newblock
\showDOI{%
\url{https://doi.org/10.1016/j.aml.2003.11.006}}


\bibitem[\protect\citeauthoryear{Liu, Yang, Li, and Zhou}{Liu
  et~al\mbox{.}}{2014}]%
        {liu2014recursive}
\bibfield{author}{\bibinfo{person}{Shujie Liu}, \bibinfo{person}{Nan Yang},
  \bibinfo{person}{Mu Li}, {and} \bibinfo{person}{Ming Zhou}.}
  \bibinfo{year}{2014}\natexlab{}.
\newblock \showarticletitle{A Recursive Recurrent Neural Network for
  Statistical Machine Translation}.
\newblock \bibinfo{journal}{{\em 52nd Annual Meeting of the Association for
  Computational Linguistics, ACL 2014 - Proceedings of the Conference\/}}
  \bibinfo{volume}{1}, \bibinfo{pages}{1491--1500}.
\newblock
\showDOI{%
\url{https://doi.org/10.3115/v1/P14-1140}}


\bibitem[\protect\citeauthoryear{{Moriarty} and {Miikkulainen}}{{Moriarty} and
  {Miikkulainen}}{1997}]%
        {moriarty1997forming}
\bibfield{author}{\bibinfo{person}{D.~E. {Moriarty}} {and} \bibinfo{person}{R.
  {Miikkulainen}}.} \bibinfo{year}{1997}\natexlab{}.
\newblock \showarticletitle{Forming Neural Networks Through Efficient and
  Adaptive Coevolution}.
\newblock \bibinfo{journal}{{\em Evolutionary Computation\/}}
  \bibinfo{volume}{5}, \bibinfo{number}{4} (\bibinfo{year}{1997}),
  \bibinfo{pages}{373--399}.
\newblock
\showDOI{%
\url{https://doi.org/10.1162/evco.1997.5.4.373}}


\bibitem[\protect\citeauthoryear{Mosavi, Faghan, Ghamisi, Duan, Ardabili,
  Salwana, and Band}{Mosavi et~al\mbox{.}}{2020}]%
        {mosavi2020comprehensive}
\bibfield{author}{\bibinfo{person}{Amir Mosavi}, \bibinfo{person}{Yaser
  Faghan}, \bibinfo{person}{Pedram Ghamisi}, \bibinfo{person}{Puhong Duan},
  \bibinfo{person}{Sina Ardabili}, \bibinfo{person}{Ely Salwana}, {and}
  \bibinfo{person}{Shahab Band}.} \bibinfo{year}{2020}\natexlab{}.
\newblock \showarticletitle{Comprehensive Review of Deep Reinforcement Learning
  Methods and Applications in Economics}.
\newblock   \bibinfo{volume}{8} (\bibinfo{date}{09} \bibinfo{year}{2020}),
  \bibinfo{pages}{1640}.
\newblock
\showDOI{%
\url{https://doi.org/10.3390/math8101640}}


\bibitem[\protect\citeauthoryear{Nagendra, Podila, Ugarakhod, and
  George}{Nagendra et~al\mbox{.}}{2017}]%
        {nagendra2017comparison}
\bibfield{author}{\bibinfo{person}{Savinay Nagendra}, \bibinfo{person}{Nikhil
  Podila}, \bibinfo{person}{Rashmi Ugarakhod}, {and} \bibinfo{person}{Koshy
  George}.} \bibinfo{year}{2017}\natexlab{}.
\newblock \showarticletitle{Comparison of Reinforcement Learning algorithms
  applied to the Cart Pole problem}. \bibinfo{pages}{26--32}.
\newblock
\showDOI{%
\url{https://doi.org/10.1109/ICACCI.2017.8125811}}


\bibitem[\protect\citeauthoryear{Nguyen, Dinh~Thai, Gong, Niyato, Wang, Liang,
  and Kim}{Nguyen et~al\mbox{.}}{2019}]%
        {nguyen2019applications}
\bibfield{author}{\bibinfo{person}{Cong Nguyen}, \bibinfo{person}{Hoang
  Dinh~Thai}, \bibinfo{person}{Shimin Gong}, \bibinfo{person}{Dusit Niyato},
  \bibinfo{person}{Ping Wang}, \bibinfo{person}{Ying-Chang Liang}, {and}
  \bibinfo{person}{Dong~In Kim}.} \bibinfo{year}{2019}\natexlab{}.
\newblock \showarticletitle{Applications of Deep Reinforcement Learning in
  Communications and Networking: A Survey}.
\newblock \bibinfo{journal}{{\em IEEE Communications Surveys \& Tutorials\/}}
  \bibinfo{volume}{PP} (\bibinfo{date}{05} \bibinfo{year}{2019}),
  \bibinfo{pages}{1--1}.
\newblock
\showDOI{%
\url{https://doi.org/10.1109/COMST.2019.2916583}}


\bibitem[\protect\citeauthoryear{Olgac and Karlik}{Olgac and Karlik}{2011}]%
        {olgac2011performance}
\bibfield{author}{\bibinfo{person}{A Olgac} {and} \bibinfo{person}{Bekir
  Karlik}.} \bibinfo{year}{2011}\natexlab{}.
\newblock \showarticletitle{Performance Analysis of Various Activation
  Functions in Generalized MLP Architectures of Neural Networks}.
\newblock \bibinfo{journal}{{\em International Journal of Artificial
  Intelligence And Expert Systems\/}}  \bibinfo{volume}{1} (\bibinfo{date}{02}
  \bibinfo{year}{2011}), \bibinfo{pages}{111--122}.
\newblock


\bibitem[\protect\citeauthoryear{Otair and Walid}{Otair and Walid}{2006}]%
        {otair2006efficient}
\bibfield{author}{\bibinfo{person}{Mohammed Otair} {and}
  \bibinfo{person}{A.~Salameh Walid}.} \bibinfo{year}{2006}\natexlab{}.
\newblock \showarticletitle{Efficient Training of Backpropagation Neural
  Networks}.
\newblock \bibinfo{journal}{{\em Neural Network World\/}}  \bibinfo{volume}{16}
  (\bibinfo{date}{01} \bibinfo{year}{2006}).
\newblock


\bibitem[\protect\citeauthoryear{Peng, Chen, Singh, and Zhang}{Peng
  et~al\mbox{.}}{2018}]%
        {peng2018neat}
\bibfield{author}{\bibinfo{person}{Yiming Peng}, \bibinfo{person}{Gang Chen},
  \bibinfo{person}{Harman Singh}, {and} \bibinfo{person}{Mengjie Zhang}.}
  \bibinfo{year}{2018}\natexlab{}.
\newblock \showarticletitle{NEAT for large-scale reinforcement learning through
  evolutionary feature learning and policy gradient search}.
  \bibinfo{pages}{490--497}.
\newblock
\showDOI{%
\url{https://doi.org/10.1145/3205455.3205536}}


\bibitem[\protect\citeauthoryear{Schober, Boer, and Schwarte}{Schober
  et~al\mbox{.}}{2018}]%
        {schober2018correlation}
\bibfield{author}{\bibinfo{person}{Patrick Schober}, \bibinfo{person}{Christa
  Boer}, {and} \bibinfo{person}{Lothar Schwarte}.}
  \bibinfo{year}{2018}\natexlab{}.
\newblock \showarticletitle{Correlation Coefficients: Appropriate Use and
  Interpretation}.
\newblock \bibinfo{journal}{{\em Anesthesia \& Analgesia\/}}
  \bibinfo{volume}{126} (\bibinfo{date}{02} \bibinfo{year}{2018}),
  \bibinfo{pages}{1}.
\newblock
\showDOI{%
\url{https://doi.org/10.1213/ANE.0000000000002864}}


\bibitem[\protect\citeauthoryear{Stanley, D'Ambrosio, and Gauci}{Stanley
  et~al\mbox{.}}{2009}]%
        {stanley2009hypercube}
\bibfield{author}{\bibinfo{person}{Kenneth Stanley}, \bibinfo{person}{David
  D'Ambrosio}, {and} \bibinfo{person}{Jason Gauci}.}
  \bibinfo{year}{2009}\natexlab{}.
\newblock \showarticletitle{A Hypercube-Based Encoding for Evolving Large-Scale
  Neural Networks}.
\newblock \bibinfo{journal}{{\em Artificial life\/}}  \bibinfo{volume}{15}
  (\bibinfo{date}{02} \bibinfo{year}{2009}), \bibinfo{pages}{185--212}.
\newblock
\showDOI{%
\url{https://doi.org/10.1162/artl.2009.15.2.15202}}


\bibitem[\protect\citeauthoryear{Stanley and Miikkulainen}{Stanley and
  Miikkulainen}{2003}]%
        {stanley2003competitive}
\bibfield{author}{\bibinfo{person}{Kenneth Stanley} {and}
  \bibinfo{person}{Risto Miikkulainen}.} \bibinfo{year}{2003}\natexlab{}.
\newblock \showarticletitle{Competitive Coevolution through Evolutionary
  Complexification}.
\newblock \bibinfo{journal}{{\em Journal of Artificial Intelligence
  Research\/}}  \bibinfo{volume}{21} (\bibinfo{date}{02} \bibinfo{year}{2003}).
\newblock
\showDOI{%
\url{https://doi.org/10.1613/jair.1338}}


\bibitem[\protect\citeauthoryear{{Stanley} and {Miikkulainen}}{{Stanley} and
  {Miikkulainen}}{2002}]%
        {stanley2002evolving}
\bibfield{author}{\bibinfo{person}{K.~O. {Stanley}} {and} \bibinfo{person}{R.
  {Miikkulainen}}.} \bibinfo{year}{2002}\natexlab{}.
\newblock \showarticletitle{Evolving Neural Networks through Augmenting
  Topologies}.
\newblock \bibinfo{journal}{{\em Evolutionary Computation\/}}
  \bibinfo{volume}{10}, \bibinfo{number}{2} (\bibinfo{year}{2002}),
  \bibinfo{pages}{99--127}.
\newblock
\showDOI{%
\url{https://doi.org/10.1162/106365602320169811}}


\bibitem[\protect\citeauthoryear{Such, Madhavan, Conti, Lehman, Stanley, and
  Clune}{Such et~al\mbox{.}}{2017}]%
        {such2017deep}
\bibfield{author}{\bibinfo{person}{Felipe Such}, \bibinfo{person}{Vashisht
  Madhavan}, \bibinfo{person}{Edoardo Conti}, \bibinfo{person}{Joel Lehman},
  \bibinfo{person}{Kenneth Stanley}, {and} \bibinfo{person}{Jeff Clune}.}
  \bibinfo{year}{2017}\natexlab{}.
\newblock \showarticletitle{Deep Neuroevolution: Genetic Algorithms Are a
  Competitive Alternative for Training Deep Neural Networks for Reinforcement
  Learning}.
\newblock  (\bibinfo{date}{12} \bibinfo{year}{2017}).
\newblock


\bibitem[\protect\citeauthoryear{Thede}{Thede}{2004}]%
        {thede2004introduction}
\bibfield{author}{\bibinfo{person}{Scott Thede}.}
  \bibinfo{year}{2004}\natexlab{}.
\newblock \showarticletitle{An introduction to genetic algorithms}.
\newblock \bibinfo{journal}{{\em Journal of Computing Sciences in Colleges\/}}
  \bibinfo{volume}{20} (\bibinfo{date}{10} \bibinfo{year}{2004}).
\newblock


\bibitem[\protect\citeauthoryear{Torrado, Bontrager, Togelius, Liu, and
  Perez-Liebana}{Torrado et~al\mbox{.}}{2018}]%
        {torrado2018deep}
\bibfield{author}{\bibinfo{person}{Ruben Torrado}, \bibinfo{person}{Philip
  Bontrager}, \bibinfo{person}{Julian Togelius}, \bibinfo{person}{Jialin Liu},
  {and} \bibinfo{person}{Diego Perez-Liebana}.}
  \bibinfo{year}{2018}\natexlab{}.
\newblock \showarticletitle{Deep Reinforcement Learning for General Video Game
  AI}. \bibinfo{pages}{1--8}.
\newblock
\showDOI{%
\url{https://doi.org/10.1109/CIG.2018.8490422}}


\bibitem[\protect\citeauthoryear{Whiteson, Stone, Stanley, Miikkulainen, and
  Kohl}{Whiteson et~al\mbox{.}}{2005}]%
        {whiteson2005automatic}
\bibfield{author}{\bibinfo{person}{Shimon Whiteson}, \bibinfo{person}{Peter
  Stone}, \bibinfo{person}{Kenneth Stanley}, \bibinfo{person}{Risto
  Miikkulainen}, {and} \bibinfo{person}{Nate Kohl}.}
  \bibinfo{year}{2005}\natexlab{}.
\newblock \showarticletitle{Automatic feature selection in neuroevolution}.
  \bibinfo{pages}{1225--1232}.
\newblock
\showDOI{%
\url{https://doi.org/10.1145/1068009.1068210}}


\bibitem[\protect\citeauthoryear{Zhang, Luo, Zhanga, and Wu}{Zhang
  et~al\mbox{.}}{2018}]%
        {zhang2018recommendation}
\bibfield{author}{\bibinfo{person}{Libo Zhang}, \bibinfo{person}{Tiejian Luo},
  \bibinfo{person}{Fei Zhanga}, {and} \bibinfo{person}{Yanjun Wu}.}
  \bibinfo{year}{2018}\natexlab{}.
\newblock \showarticletitle{A Recommendation Model Based on Deep Neural
  Network}.
\newblock \bibinfo{journal}{{\em IEEE Access\/}}  \bibinfo{volume}{PP}
  (\bibinfo{date}{01} \bibinfo{year}{2018}), \bibinfo{pages}{1--1}.
\newblock
\showDOI{%
\url{https://doi.org/10.1109/ACCESS.2018.2789866}}


\end{thebibliography}

\end{document}